\documentclass[11pt]{article}
\usepackage{geometry}
\usepackage{coling2020}
\usepackage{times}
\usepackage{url}
\usepackage{latexsym}
\usepackage{microtype}

\usepackage{makecell}
\usepackage{amsmath}
\usepackage[caption=false]{subfig}
\usepackage[hidelinks]{hyperref}

\colingfinalcopy % Uncomment this line for all SemEval submissions

\title{BRUMS at SemEval-2020 Task 3: Contextualised Embeddings for Predicting the (Graded) Effect of Context in Word Similarity}

\author{Hansi Hettiarachchi$^1$, Tharindu Ranasinghe$^2$ \\

$^1$School of Computing and Digital Technology, Birmingham City University, UK \\

$^2$Research Group in Computational Linguistics, University of Wolverhampton, UK \\
{\tt hansi.hettiarachchi@mail.bcu.ac.uk } \\
{\tt t.d.ranasinghehettiarachchige@wlv.ac.uk } } 

\date{}

\begin{document}
\maketitle
\begin{abstract}
This paper presents the team \textit{BRUMS} submission to SemEval-2020 Task 3: Graded Word Similarity in Context. The system utilises state-of-the-art contextualised word embeddings, which have some task-specific adaptations, including stacked embeddings and average embeddings. Overall, the approach achieves good evaluation scores across all the languages, while maintaining simplicity. Following the final rankings, our approach is ranked within the top 5 solutions of each language while preserving the 1\textsuperscript{st} position of Finnish subtask 2.
\end{abstract}

\section{Introduction}
\label{sec:introduction}

% The following footnote without marker is needed for the camera-ready
% version of the paper.
% Comment out the instructions (first text) and uncomment the 8 lines
% under "final paper" for your variant of English.
% 
\blfootnote{
    %
    % for review submission
    %
    % \hspace{-0.65cm}  % space normally used by the marker
    % Place licence statement here for the camera-ready version. 
    % See
    % Section~\ref{licence} of the instructions for preparing a
    % manuscript.
    %
    % % final paper: en-uk version 
    %
    \hspace{-0.65cm}  % space normally used by the marker
    This work is licensed under a Creative Commons 
    Attribution 4.0 International Licence.
    Licence details:
    \url{http://creativecommons.org/licenses/by/4.0/}.
    % 
    % % final paper: en-us version 
    %
    % \hspace{-0.65cm}  % space normally used by the marker
    % This work is licensed under a Creative Commons 
    % Attribution 4.0 International License.
    % License details:
    % \url{http://creativecommons.org/licenses/by/4.0/}.
}

In natural language, meaning of a word has an influence by its surrounding or context words. This impact is mainly driven by the associated linguistic and cognitive phenomena \cite{armendariz-EtAl:2020:LREC}. Following these two phenomena, it was found that word meanings have a continuous nature in addition to the commonly known discrete nature. 

From the cognitive perspective, word meanings can be assigned based on the conceptual structures in human mind \cite{evans2006cognitive,gardenfors2014geometry}. Therefore, meaning of a word can be varied according to the mental state of the reader which is triggered with the contact of context words. Thus, word meanings are not just limited to a discrete nature. From the linguistic perspective, context words can modify the meaning of a word by contextual selection and contextual modulation \cite{cruse1986lexical}. Contextual selection identifies the meanings of polysemous words (i.e. words with multiple senses) such as \textit{`cell'}, \textit{`bank'}, etc. using the context words. In this case, the most appropriate meaning will be picked from a set of discrete senses. For an example, phrase \textit{`prison cell'} implies that the word \textit{`cell'} refers a room. Contextual modulation modifies the meanings of single sense words by highlighting their characteristics. Thus, unlike the contextual selection, modulation makes a continuous effect on the meaning and it is widely available, because majority of the words are general to a certain extent. For an example, \textit{`butter'} is a single sense word. But, if it appears in the phrase \textit{`poured the butter'}, context words reveal that the mentioned butter is in liquid state. Focusing on the above-mentioned facts, it is important to consider both discrete and continuous effects while predicting the meaning of words in natural language text.

% Contextual selection identifies the meanings of ambiguous words such as \textit{`cell'}, \textit{`bank'}, etc. using the context words. In this case, the most appropriate meaning will be picked from a set of discrete senses. For an example, phrase; \textit{`prison cell'} implies that the word; \textit{`cell'} ers to a room. Contextual modulation modifies the meanings of single sense words by highlighting their characteristics. Thus, unlike the contextual selection, modulation has a continuous effect on the meaning and it is widely available, because majority of the words are general to a certain extent. For an example, \textit{`butter'} is a single sense word. But, if it appears in the phrase; \textit{`poured the butter'}, context words reveal that the mentioned butter is in liquid state. Focusing on the above-mentioned facts, it is important to consider both discrete and continuous effects while predicting the meaning of words in natural language text. 

Even though there is a continuous effect on meaning, majority of the previous research were only focused on the discrete effect. Considering this limitation, SemEval-2020 Task 3 was designed by targeting the continuous (graded) effects of context. With the involvement of continuous effects, the goal of this task is predicting the effect of context in human perception of similarity. For an example, given the phrases: 

\begin{quote}
    \textit{`Her prison \textbf{cell} was almost an improvement over her \textbf{room} at the last hostel.'}
\end{quote} 

\begin{quote}
    \textit{`His job as a biologist didn't leave much \textbf{room} for a personal life. He knew much more about human \textbf{cells} than about human feelings.'}
\end{quote}

this task needs to predict that words \textit{`room'} and \textit{`cell'} have more similar meaning in phrase 1 compared to phrase 2.

% \textit{`Her prison \textbf{cell} was almost an improvement over her \textbf{room} at the last hostel.'} and \textit{`His job as a biologist didn't leave much \textbf{room} for a personal life. He knew much more about human \textbf{cells} than about human feelings.'}, this task need to predict that words; \textit{`room'} and \textit{`cell'} have more similar meaning in phrase 1 compared to phrase 2.

% As participants of SemEval-2020 Task 3, we experimented the impact of contextualised word embeddings extracted by different architectures and learning methods on predicting the (graded) effect of context in word similarity. Rather than focusing on the embeddings taken with default settings, we evaluated stacked embeddings, different parameter settings, average embeddings and improved known data rate in this research. 

As participants of SemEval-2020 Task 3, we experimented the impact of contextualised word embeddings extracted by different architectures and learning methods on predicting the (graded) effect of context in word similarity. Rather than focusing on the embeddings taken with default settings, we evaluated stacked embeddings, different parameter settings and average embeddings. Further, we experimented the embeddings generated with improved known data rate. In this approach, model unknown words in the data set are replaced with model known words to support more effective embedding generation. The rest of this paper is organised as follows. Section \ref{sec:task-data-set} describes the task including its subtasks and data sets. Available methods related to this research are summarised under Section \ref{sec:related-work} and our approaches are described in Section \ref{sec:methodology}. Following these, results are mentioned in Section \ref{sec:results} and paper is concluded with Section \ref{sec:conclusion}. 

% Since we are using pre-trained word embedding models, there exist model unknown words in the data set and we replaced such words with known words to have high rate of known words to support embedding generation. 

\section{Task description and Data sets} \label{sec:task-data-set}
SemEval-2020 Task 3 \cite{armendariz-etal-2020-semeval} is focused on predicting the (graded) effect of context in word similarity. There were two unsupervised subtasks associated with this shared task as follows; 
\begin{itemize}
    \item {\bf Subtask 1 - Predicting Changes}: Predicting the degree and direction of change in similarity of same word pair within two different contexts ( This task targets the ability to identify continuous effects on meaning or effects made by context in human perception of similarity )
    
    \item {\bf Subtask 2 - Predicting Ratings}: Predicting the similarity of same word pair within two different contexts ( This task is more similar to the traditional task which evaluates the similarity of words based on their contexts )
\end{itemize}

As the evaluation data set, CoSimLex \cite{armendariz-EtAl:2020:LREC} was used. This data set is newly created using human annotators, with the focus on graded effect on word similarity. As the first version, 340 English pairs, 112 Croatian pairs, 111 Slovenian pairs and 24 Finnish pairs were released without human annotated scores to use with the evaluation phase of SemEval-2020 Task 3. Among them, randomly selected 10 English pairs, 5 Croatian pairs and 5 Slovenian pairs were released including the human annotated scores as the practice data for participants to evaluate their models.

\section{Related Work} \label{sec:related-work}
Effect of context was considered by previous research to predict the meaning of words and their similarity. In order to include wider context knowledge, Huang et al. \shortcite{huang2012improving} introduced a neural network-based language model which considers both local and global contexts to learn word representations. As local context, sequence of words is considered and as global context, corresponding document is considered. To capture the multiple senses of a word, they suggested to cluster learned context word vectors into different meaning groups.

In later research, there was a tendency to focus on sense embeddings. Sense embeddings are vectors generated to represent different senses of words. Chen et al. \shortcite{chen2014unified} suggested to use words in WordNet \cite{miller1995wordnet} gloss contents to produce sense embeddings. To obtain vector representations of words, they used Skip-gram model \cite{mikolov2013efficient}. For each polysemous word, a sense was picked by comparing the similarity between its context vector and available sense vectors. Given a sentence, average of its word vectors was computed as the context vector.  Similar approach was suggested by another recent research using BERT embeddings \cite{devlin2018bert} to incorporate more contextual features \cite{loureiro2019liaad}. In this approach, sense embedding generation was initiated using SemCor corpus \cite{miller1994using}. Using the sentences available in the corpus for each sense, BERT embeddings were generated and average of those embeddings were taken as the sense embedding. To improve the sense coverage, WordNet's ontology and glosses also considered. Given a sentence, generated BERT embedding corresponding to a polysemous word was compared with sense embeddings to disambiguate it. 

Using meaning groups and sense embeddings, above-mentioned approaches are only looking for discrete word meanings with the effect of context and we could not find any approach which considers both discrete and continuous effects. As we are aware of CoSimLex is the first data set which is annotated by considering the continuous effect of context and SemEval-2020 Task 3 is the first shared task which is focused on predictions based on this effect. However, most of the available research are based on prediction-based word embeddings and among them contextualised word embeddings are found to be more capable in extracting word meanings based on the context. 

\section{Methodology} \label{sec:methodology}
This section describes the different approaches used to predict the (graded) effect of context in word similarity. Since the similarity between a word pair is measured using the cosine similarity between word vectors, we experimented different methods to generate vector representations based on recently published contextualised word embedding models as further described in Sections; \ref{ssec:embedding-models} - \ref{ssec:named-entities}. All the implementations are done in Python \footnote{The code is available on \url{https://github.com/HHansi/Semeval-2020-Task3}}.

\subsection{Contextualised Embeddings}
\label{ssec:embedding-models}
Unlike the classic word embeddings; Word2Vec \cite{mikolov2013efficient}, GloVe \cite{pennington2014glove}, fastText \cite{bojanowski2017enriching}, etc., contextualised word embeddings capture the variations of word meanings based on their context. Therefore, they were successfully applied to wide range of NLP tasks such as text classification \cite{10.1007/978-3-030-32381-3_16,ranasinghe2019brums}, question answering \cite{devlin2018bert,alloatti-etal-2019-real} and machine translation \cite{imamura-sumita-2019-recycling,Zhu2020Incorporating} with improved performance. 

Among the various contextualised embedding models available, we used ELMo \cite{peters2018deep}, Flair \cite{akbik2018contextual}, BERT \cite{devlin2018bert}, Transformer-XL \cite{dai2019transformer} and XLNet \cite{yang2019xlnet} for this task in order to evaluate the impact by different architectures and learning methods on contextual word similarity prediction. Both ELMo and Flair are based on bidirectional LSTM (Long Short-Term Memory) architecture \cite{sundermeyer2012lstm} and other models are based on bidirectional Transformer architecture \cite{devlin2018bert}. ELMo is learned on sequence of tokens and Flair is learned on sequence of characters. The Transformer architecture introduced by BERT was extended as Transformer-XL by enabling the learning dependencies beyond fixed length contexts. XLNet is a further improved version of Transformer-XL considering the advantages in autoregressive and autoencoding methods. All these Transformer-based models are learned on sequence of tokens. 

This research is supported by available pretrained embedding models. Fixed size word vectors are generated for each of the target words using their given context and pretrained model weights. The experimented models including their language coverage are summarised in Table \ref{table:models}. We used the implementations by FLAIR \cite{akbik2019flair}\footnote{Git repository of flair is available on \url{https://github.com/flairNLP/flair}} and Hugging Face \footnote{All the models supported by Hugging Face can be found on \url{https://huggingface.co/models}} for embedding generation. More details about the models including layers, parameters and training corpora are available with FLAIR and Hugging Face documentation. 

\begin{table}[t!]
\begin{center}
\begin{tabular}{|l|l|l|l|l|}
\hline \bf Model & \bf en & \bf hr & \bf sl  & \bf fi \\ \hline
ELMo:large & x & - & - & - \\
\hline
Flair:multi-X & x & x & x & x \\
\hline
Flair:hr-X & - & x & - & - \\
\hline
Flair:sl-X & - & - & x & - \\
\hline
Flair:sl-v0-x & - & - & x & - \\
\hline
Flair:fi-X & - & - & - & x \\
\hline
BERT:large-cased & x & - & - & - \\
\hline
BERT:large-uncased & x & - & - & - \\ 
\hline
BERT:base-multilingual-cased & x & x & x & x \\
\hline
BERT:base-multilingual-uncased & x & x & x & x \\
\hline
BERT:base-finnish-cased-v1 & - & - & - & x \\
\hline
BERT:base-finnish-uncased-v1 & - & - & - & x \\
\hline
Transformer-XL:wt103 & x & - & - & - \\
\hline
XLNet:large-cased & x & - & - & - \\
\hline
\end{tabular}
\end{center}
\caption{ Embedding models used by this research and their language coverage based on the languages; English(en), Croatian(hr), Slovenian(sl) and Finnish(fi) specific to this task} \label{table:models}
\end{table}

\subsection{Stacked Embeddings} \label{ssec:stacked-embedding}
Secondly, we tried out the stacked embeddings generated using above-mentioned (Section \ref{ssec:embedding-models}) contextualised word embeddings. Stacked embeddings combine each vector by concatenating them to form the final vector \cite{akbik2018contextual} as shown in Equation \ref{eq:stacked-embedding}. $v_{i}^{stk}$ represents the final or stacked word vector corresponding to the word $i$ and $v_{i}^{model_{m}}$ represents the vector obtained by using the embedding model $m$. Combining word embeddings from different learning methods allows to combine their characteristics together. For this research, we experimented the stacked embeddings by combining up to three models only. 

\begin{align} \label{eq:stacked-embedding}
    v_{i}^{stk} &= \begin{bmatrix}
       v_{i}^{model_{1}} \\
       v_{i}^{model_{2}} \\
       \vdots \\
       v_{i}^{model_{m}}
     \end{bmatrix}
\end{align}

% \subsection{Parameter Optimisation with BERT}
\subsection{Parameter Settings with BERT} \label{ssec:param-optimisation}
As the next approach, we experimented the effectiveness of BERT embeddings using different parameter settings for embedding extraction layers, sub-token selection and scalar mix. Further, we tested the impact by average embeddings also. BERT model was selected for this experiment, because we could obtain good results for all languages using it. 

Embedding extraction layers indicate from which layers of the learned model, weights need to be taken to represent the word vectors. For all the above experiments mentioned in Section \ref{ssec:embedding-models} and \ref{ssec:stacked-embedding}, we used the concatenation of last four layers, because it was found as the best approach to represent the features in underlying text \cite{devlin2018bert}. 

Since Transformer-based models use sub-tokens to get the embeddings, we analysed the impact by different sub-token selection techniques; first, last, concatenation of first and last (first-last) and mean on predicting the word similarity. Only the first sub-token was used for above experiments. 

Scalar mix allows the computation of parameterised scalar mixture of the defined layers \cite{tenney2019bert}. This technique was found to be useful, because the best performing layer of a Transformer model could vary depending on the task and it is unclear to the user. Further, Liu et al. \shortcite{liu2019linguistic} found that scalar mix of Transformer layers has the ability in outperforming the best individual layers. For above experiments, no scalar mix was used. 

{\bf Average Embeddings}: As average embeddings, we considered the average of weights in different layers in order to combine the information learned by them together. For word $i$, by considering the last $k$ layers, average embedding $v_{i}^{avg}$ is calculated by following the Equation \ref{eq:average-embeddings}. Weights in the last layer are represented by the vector $v_{i}^{-1}$. 

\begin{equation} \label{eq:average-embeddings}
    v_{i}^{avg} = \frac{v_{i}^{-k} + ... + v_{i}^{-1}}{k}
\end{equation}

% \subsection{Named Entities for Unknown Tokens}
\subsection{Improved Known Data Rate}
\label{ssec:named-entities}
Since we used pretrained embedding models for this research, there were tokens such as person names, organisations, locations, etc. in the data set which are unknown to the vocabulary of the pretrained model. By replacing them with some known tokens, we can provide more familiar data to the model during the embedding generation. To convert these tokens into a known form automatically, we used Named Entity Recognition (NER) \cite{nadeau2007survey}. Named entities which are identified using the models available with spaCy \footnote{More details about spaCy are available on \url{https://spacy.io/}} were used to replace the unknown tokens. For an example, \textit{`...underground in the late 1960s, \textbf{Sihanouk} had to make concessions...'} is converted as \textit{`...underground in the late 1960s, \textbf{person} had to make concessions...'}. But, considering the limited availability of NER models, we conducted this experiment only on English data set.

\section{Results} \label{sec:results}
This section contains the results obtained for both subtasks using the approaches described in Section \ref{sec:methodology}. We evaluated the suggested methods using practice data and used the best methods for submissions. The top three scores obtained by submissions on both tasks for all four languages and baseline scores are summarised under Section \ref{ssec:results-subtask1} and \ref{ssec:results-subtask2}. 

For all the reported Transformer-based models, as the default parameter setting, concatenation of last four layers using first sub-token without scalar mix is used. Any parameter change except the default setting is mentioned within the brackets after the model name. Stacked models are mentioned using $+$ symbol (e.g. $model_{1} + model_{2}$) and phrase \textit{with NE} is appended to model name if improved known data rate is used.  

As the baseline model for both tasks, the multilingual BERT model released by Gary Lai \footnote{Git repository of baseline embedding model is available on \url{https://github.com/imgarylai/bert-embedding}} was used as informed by the task organisers. This model was trained using an uncased multilingual data set extracted from Wikipedia.

\subsection{Subtask 1} \label{ssec:results-subtask1}
To evaluate the subtask 1 results, Pearson correlation between the model predicted values and gold standards was measured. Average values of scores produced by human annotators are used as gold standards. Since this task is to measure the change in similarity between two contexts, the sign of results is also an important measure, because it indicates the direction of change. Therefore, the uncentered variation of the Pearson correlation which is calculated using the standard deviation from zero was used. 

The top three results we obtained with the baseline results for English, Croatian, Slovenian and Finnish are shown in Table \ref{tab:results-1}.  According to the results, our approaches could outperform the baseline in all languages except Finnish. English and Croatian used the average embeddings to obtain the best score while Slovenian used the stacked embeddings. 

%[hbt]
\begin{table}[!ht]
    \centering
    % \begin{tabular}{p{0.45\textwidth}p{0.45\textwidth}}

    \subfloat[Final evaluation results-English]{
        \begin{tabular}{|p{0.36\textwidth}|p{0.06\textwidth}|}
        \cline{1-2}
        {\bf Model} & {\bf Score} \\
        \cline{1-2}
        \makecell[l]{baseline} & 0.713 \\
        \cline{1-2}
        \makecell[l]{BERT:large-cased+Transformer-XL\\:wt103} & 0.684 \\
        \cline{1-2}
        \makecell[l]{BERT:large-cased(k=14)} & \bf 0.754  \\
        \cline{1-2}
        \makecell[l]{BERT:large-cased(k=14):with NE} & 0.753 \\
        \cline{1-2}
        \end{tabular}
    }
   \hfill
    \subfloat[Final evaluation results-Croatian]{
        \begin{tabular}{|p{0.36\textwidth}|p{0.06\textwidth}|}
        \cline{1-2}
        {\bf Model} & {\bf Score} \\
        \cline{1-2}
        \makecell[l]{baseline} & 0.587\\
        \cline{1-2}
        \makecell[l]{BERT:base-multilingual-uncased} & 0.651 \\
        \cline{1-2}
        \makecell[l]{BERT:base-multilingual-cased(first-\\last)} & \bf 0.664  \\
        \cline{1-2}
        \makecell[l]{BERT:base-multilingual-cased(k=4)} & \bf 0.664 \\
        \cline{1-2}
        \end{tabular}
    }
    % \end{tabular}
    
    % \begin{tabular}{p{0.45\textwidth}p{0.45\textwidth}}
    \subfloat[Final evaluation results-Slovenian]{
        \begin{tabular}{|p{0.36\textwidth}|p{0.06\textwidth}|}
        \cline{1-2}
        {\bf Model} & {\bf Score} \\
        \cline{1-2}
        \makecell[l]{baseline} & 0.603\\
        \cline{1-2}
        \makecell[l]{Flair:sl-forward+Flair:sl-backward+\\BERT:base-multilingual-uncased} & \bf 0.648 \\
        \cline{1-2}
        \makecell[l]{BERT:base-multilingual-cased(k=4)} & 0.608 \\
        \cline{1-2}
        \makecell[l]{BERT:base-multilingual-cased(k=6)} & 0.621 \\
        \cline{1-2}
        \end{tabular}
    }
    % &
    \hfill
    \subfloat[Final evaluation results-Finnish]{
        \begin{tabular}{|p{0.36\textwidth}|p{0.06\textwidth}|}
        \cline{1-2}
        {\bf Model} & {\bf Score} \\
        \cline{1-2}
        \makecell[l]{baseline} & \bf 0.671\\
        \cline{1-2}
        \makecell[l]{BERT:finnish-cased-v1} & 0.642 \\
        \cline{1-2}
        \makecell[l]{BERT:finnish-uncased-v1} & 0.594 \\
        \cline{1-2}
        \makecell[l]{BERT:finnish-cased-v1(k=6)} & 0.626 \\
        \cline{1-2}
        \end{tabular}
    }
    % \end{tabular}
    \caption{Subtask 1 results on final evaluation}
    \label{tab:results-1}
\end{table}

\subsection{Subtask 2} \label{ssec:results-subtask2}
To evaluate the subtask 2 results, harmonic mean of Pearson and Spearman correlations between model predicted values and gold standards was used \cite{camacho2017semeval}. 

The top three resutls we obtained with the baseline results for English, Croatian, Slovenian and Finnish are shown in Table \ref{tab:results-2}. Based on the results, our approaches outperformed the baseline in all languages. The best score in each language was obtained using BERT embeddings which used first-last sub-token with scalar mix. 

% For majority; English, Croatian and Finnish, the best score was obtained by the BERT embeddings which used first-last sub-token with scalar mix. Exceptionally, best score for Slovenian was obtained using the average embeddings. 

% [hbt]
\begin{table}[!ht]
    \centering
    % \begin{tabular}{p{0.45\textwidth}p{0.45\textwidth}}

    \subfloat[Final evaluation results-English]{
        \begin{tabular}{|p{0.36\textwidth}|p{0.06\textwidth}|}
        \cline{1-2}
        {\bf Model} & {\bf Score} \\
        \cline{1-2}
        \makecell[l]{baseline} & 0.573 \\
        \cline{1-2}
        \makecell[l]{BERT:base-multilingual-uncased} & 0.570 \\
        \cline{1-2}
        \makecell[l]{BERT:large-cased(first-last,scalar-\\mix)} & \bf 0.715  \\
        \cline{1-2}
        \makecell[l]{BERT:large-cased(first-last,scalar-\\mix):with NE} & 0.713 \\
        \cline{1-2}
        \end{tabular}
    }
   \hfill
    \subfloat[Final evaluation results-Croatian]{
        \begin{tabular}{|p{0.36\textwidth}|p{0.06\textwidth}|}
        \cline{1-2}
        {\bf Model} & {\bf Score} \\
        \cline{1-2}
        \makecell[l]{baseline} & 0.402\\
        \cline{1-2}
        \makecell[l]{BERT:base-multilingual-cased} & 0.482 \\
        \cline{1-2}
        \makecell[l]{BERT:base-multilingual-uncased\\(first-last)} & 0.528  \\
        \cline{1-2}
        \makecell[l]{BERT:base-multilingual-uncased\\(first-last,scalar-mix)} & \bf 0.545 \\
        \cline{1-2}
        \end{tabular}
    }
    % \end{tabular}
    
    % \begin{tabular}{p{0.45\textwidth}p{0.45\textwidth}}
    \subfloat[Final evaluation results-Slovenian]{
        \begin{tabular}{|p{0.36\textwidth}|p{0.06\textwidth}|}
        \cline{1-2}
        {\bf Model} & {\bf Score} \\
        \cline{1-2}
        \makecell[l]{baseline} & 0.516\\
        \cline{1-2}
        \makecell[l]{BERT:base-multilingual-cased} & 0.524 \\
        \cline{1-2}
        \makecell[l]{BERT:base-multilingual-cased(k=4)} & 0.524  \\
        \cline{1-2}
        \makecell[l]{BERT:base-multilingual-uncased(first-\\last,scalar-mix)} & \bf 0.573 \\
        \cline{1-2}
        \end{tabular}
    }
    % &
    \hfill
    \subfloat[Final evaluation results-Finnish]{
        \begin{tabular}{|p{0.36\textwidth}|p{0.06\textwidth}|}
        \cline{1-2}
        {\bf Model} & {\bf Score} \\
        \cline{1-2}
        \makecell[l]{baseline} & 0.289\\
        \cline{1-2}
        \makecell[l]{BERT:finnish-cased-v1} & 0.644 \\
        \cline{1-2}
        \makecell[l]{BERT:finnish-uncased-v1} & 0.636  \\
        \cline{1-2}
        \makecell[l]{BERT:finnish-uncased-v1(first-last,\\scalar-mix)} & \bf 0.645 \\
        \cline{1-2}
        \end{tabular}
    }
    % \end{tabular}
    \caption{Subtask 2 results on final evaluation}
    \label{tab:results-2}
\end{table}

\section{Conclusion} \label{sec:conclusion}
In this paper, we presented different approaches used for SemEval-2020 Task 3: Graded Word Similarity in Context. We mainly evaluated the recent contextualised word embedding models with different embedding generation techniques. Depending on the differences between languages, we could not find any universal approach suitable for all languages to predict the effect of context in word similarity. For subtask 1: predicting changes, best results were obtained using average embeddings and stacked embeddings. It concludes that by combining weights in different layers of same model or different models, degree and direction of change in similarity can be predicted more accurately. For subtask 2: predicting ratings, best results were obtained by BERT embeddings which are generated using first-last sub-token with scalar mix. Different pretrained models were performed well on each language. It concludes that the capability of BERT embeddings on predicting the similarity of words based on their contexts can be further improved using appropriate parameter settings. 

% For subtask 2: predicting ratings, best results were obtained by BERT embeddings which are generated using first-last sub-token with scalar mix on different pretrained models. It concludes that the capability of BERT embeddings on predicting the similarity of words based on their contexts can be further improved using appropriate parameter settings. 

As future directions of this research, we hope to experiment the impact by learning and fine tuning options of contextual embedding models on graded effect of context in word similarity. 

% \clearpage

\bibliographystyle{coling}
\bibliography{semeval2020}

\end{document}